\documentclass[10pt,conference,letterpaper]{IEEEtran}
\usepackage{subcaption}
\usepackage{cite}
\usepackage{amsmath,amssymb,amsfonts}
\usepackage{algorithmic}
\usepackage{graphicx}
\usepackage{textcomp}
\usepackage{xcolor}
\usepackage{soul}
\usepackage{enumerate}
\usepackage{etoolbox}
\usepackage[font=small]{caption}
\def\BibTeX{{\rm B\kern-.05em{\sc i\kern-.025em b}\kern-.08em
   T\kern-.1667em\lower.7ex\hbox{E}\kern-.125emX}}

\begin{document}

\title{{Data-Driven Spectrum Demand Prediction: A Spatio-Temporal Framework with Transfer Learning}}

\author{
Amin Farajzadeh\textsuperscript{*1}, Hongzhao Zheng\textsuperscript{1}, Sarah Dumoulin\textsuperscript{2}, Trevor Ha\textsuperscript{2},\\ Halim Yanikomeroglu\textsuperscript{1}, and Amir Ghasemi\textsuperscript{2} \\ \textsuperscript{1}\textit{Carleton University, Ottawa, ON, Canada}\\
\textsuperscript{2}\textit{Communications Research Centre (CRC) Canada, Ottawa, ON, Canada}\\
\textsuperscript{*}Corresponding Email: aminfarajzadeh@cmail.carleton.ca
}

\makeatletter

\maketitle

\begin{abstract}
Accurate spectrum demand prediction is crucial for informed spectrum allocation, effective regulatory planning, and fostering sustainable growth in modern wireless communication networks. It supports governmental efforts, particularly those led by the international telecommunication union (ITU), to establish fair spectrum allocation policies, improve auction mechanisms, and meet the requirements of emerging technologies such as advanced 5G, forthcoming 6G, and the internet of things (IoT). This paper presents an effective spatio-temporal prediction framework that leverages crowdsourced user-side key performance indicators (KPIs) and regulatory datasets to model and forecast spectrum demand. The proposed methodology achieves superior prediction accuracy and cross-regional generalizability by incorporating advanced feature engineering, comprehensive correlation analysis, and transfer learning techniques. Unlike traditional ITU models, which are often constrained by arbitrary inputs and unrealistic assumptions, this approach exploits granular, data-driven insights to account for spatial and temporal variations in spectrum utilization. Comparative evaluations against ITU estimates, as the benchmark, underscore our framework's capability to deliver more realistic and actionable predictions. Experimental results validate the efficacy of our methodology, highlighting its potential as a robust approach for policymakers and regulatory bodies to enhance spectrum management and planning. \end{abstract}
\section{Introduction}
The exponential growth of wireless communication systems has created an unprecedented demand for radio spectrum, driven particularly by the proliferation of advanced 5G networks, the massive-scale internet of things (IoT), and the emergence of next-generation 6G concepts, such as integrated non-terrestrial networks (NTN). This surge in demand presents a critical challenge, especially for governments and regulatory bodies tasked with ensuring equitable spectrum allocation and optimizing resource utilization~\cite{intro-spectrum}. Therefore, accurate, data-driven prediction of spectrum demand is essential for informed decision-making at both national and regional levels~\cite{intro-spectrum-2}.

The international telecommunication union (ITU) has long played a pivotal role in addressing spectrum management challenges through its standardized models. These models develop global benchmarks for high- and low-density environments, serving as a foundational reference for policymakers~\cite{itu}. However, the ITU models often rely on static assumptions, such as uniform traffic patterns and idealized user distributions, which do not account for the nuanced variations found in real-world deployments. Consequently, these estimates frequently overstate or understate actual spectrum requirements, which hampers their relevance to dynamic and region-specific scenarios~\cite{unreal-itu}.

To address these limitations, this study introduces a novel, data-driven approach specifically designed for regulatory bodies and government-level spectrum management. By leveraging crowdsourced datasets that encompass granular network performance metrics, such as normalized number of connections, sum throughput, latency, and signal strength, we aim to develop predictive models that enhance accuracy and scalability. Unlike traditional methods, our approach caters to both spatial and temporal variations, enabling policymakers to make more informed decisions regarding spectrum allocation.

A cornerstone of this research is the identification and application of user-centric, data-driven key performance indicators (KPIs) derived from crowdsourced datasets. While traditional methodologies often rely on cell-side KPIs from service providers~\cite{unreal-itu-2}, our approach prioritizes user-side metrics that capture nuanced aspects of network performance. These metrics reflect user behavior patterns, application usage trends, and device-specific traffic characteristics. By analyzing the correlations between these KPIs and spectrum demand, we uncover deeper insights into the key drivers of spectrum utilization, paving the way for more accurate and actionable spectrum management strategies.

Our methodology integrates spatio-temporal data preparation, correlation analysis with current and lagged KPIs, and predictive modeling, augmented with transfer learning techniques. Transfer learning enhances prediction accuracy by utilizing knowledge gained in one region to improve predictions in others, thereby addressing the challenges of cross-regional generalizability. This proposed framework is particularly well-suited for regulatory bodies, as it emphasizes broad-scale insights over the localized, operational perspectives typically emphasized by service providers.
\begin{figure*}[!t]
    \centering
    \includegraphics[width=0.8\linewidth]{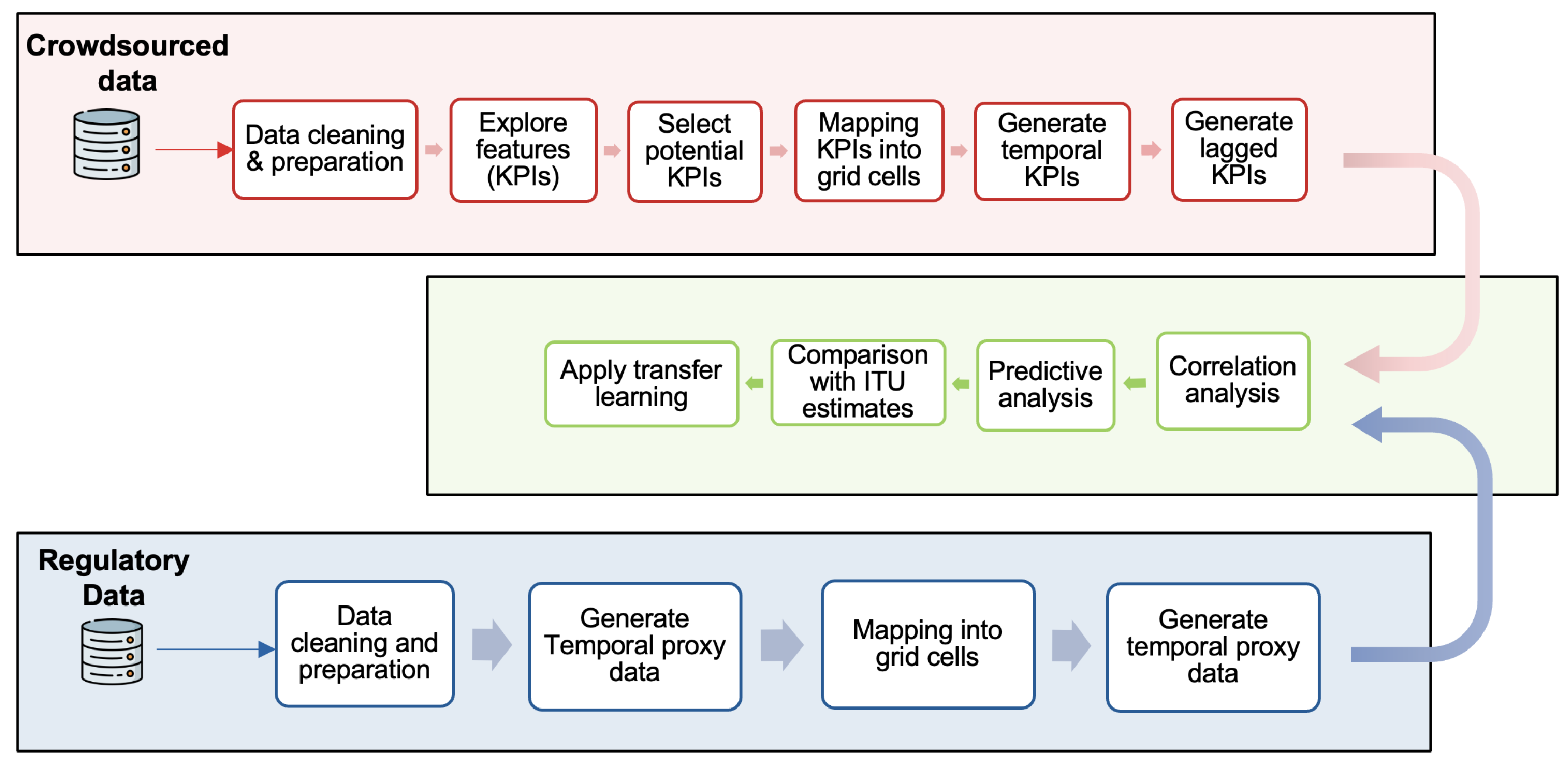}
    \caption{Overview of the proposed spatio-temporal spectrum demand prediction methodology.}
    \label{fig1}
\end{figure*}

This paper aims to address the following research questions:
\begin{itemize}
    \item Can the integration of crowdsourced user-side KPIs with regulatory datasets, within a comprehensive spatio-temporal analysis framework, effectively predict spectrum demand at both regional and national levels?
\item Is it necessary to pair each current KPI with its lagged version to capture spectrum demand more accurately, given that it usually does not respond instantly to changes in network conditions?
\item How can transfer learning techniques enhance the accuracy, generalizability, and spatial-temporal sensitivity of spectrum demand predictions across diverse geographic regions?
\item How do the predictions generated by our proposed methodology, which incorporates both explainable (white-box) and non-explainable (black-box) models, compare with traditional ITU estimates, and what are the broader implications for policy formulation and regulatory practices?
\end{itemize} 
\noindent
By addressing these questions, the study not only highlights the shortcomings of existing ITU models but also demonstrates the potential of data-driven methodologies to transform spectrum management. The findings hold significant implications for governments and regulatory bodies, offering actionable insights for equitable spectrum allocation and efficient spectrum management strategies.
\section{Related Work}

In recent years, predicting spectrum demand has attracted significant attention, leading to the development of various methodologies aimed at enhancing both accuracy and applicability. Early approaches, such as those proposed by the ITU~\cite{itu},~\cite{itu2}, primarily relied on statistical modeling to establish macro-level benchmarks for spectrum allocation on a global scale. Expanding on this foundation, researchers in~\cite{stat_itu_Malaysia} incorporated trend data from both ITU models alongside site utilization factors and capacity estimates. This approach assesses mobile broadband spectrum demand by merging statistical analyses with actual operational metrics.

Furthermore, another study~\cite{lit_queuing} employed analytical queuing models in conjunction with empirical traffic behavior observations to evaluate the spectrum requirements specific to wireless systems. More recently, the methodology introduced in~\cite{lit_indonesia} and~\cite{lit_turkey} combined market research with an advanced forecasting model to predict future spectrum needs. Their structured forecasting process begins with statistical traffic load calculations, proceeds to capacity determination for quality of service, and ultimately culminates in a spectrum estimation that factors in spectral efficiency.

Despite the valuable insights provided by these models, they often fail to capture the fine-grained, context-specific nature of spectrum demand. Their dependence on static assumptions and idealized conditions limits their ability to address regional and temporal variations in spectrum usage—an essential consideration for the evolving demands of next-generation networks~\cite{crc-brown}.

To address these limitations, recent research has increasingly gravitated toward data-driven methods that account for both spatial and temporal dynamics, leveraging advanced machine learning models. For instance, the work in~\cite{lit-deepL} introduced a deep learning framework focused on predicting crowdsourced service availability by analyzing historical spatio-temporal traces of mobile services. Their two-stage model first clusters services by geographic region and then frames the availability prediction as a classification task, effectively capturing the inherent spatial and temporal patterns.

In a similar fashion, the authors in~\cite{lit-deepair} developed ``DeepAir'', a hierarchical deep learning framework designed for real-time forecasting of KPIs in cellular networks. By utilizing stacked long short-term memory (LSTM) networks, their approach successfully models instantaneous, periodic, and seasonal patterns within KPI data, thus significantly enhancing real-time prediction accuracy in complex cellular environments.

Building on data-driven techniques, the methodology presented in~\cite{lit-geo-mohammed} integrates machine learning with geospatial analytics to estimate spectrum demand and identify critical drivers within the mobile broadband domain. Although their analysis is geographically oriented and lacks temporal considerations, the case studies conducted in Canada showcase the effectiveness of this approach.

Furthermore, the realm of transfer learning has shown considerable promise in situations characterized by limited data availability. As noted in~\cite{lit-transferL}, applying transfer learning to the forecasting of intermittent time series has proven effective in reducing forecasting errors, especially for short and medium-length time series.

Despite these advancements, a notable gap remains: the integration of crowdsourced user-side KPIs with regulatory datasets to conduct a comprehensive, joint spatial and temporal analysis. Additionally, there is a pressing need to explore explainable (``white-box'') and non-explainable (``black-box'') models, alongside transfer learning techniques, to predict future spectrum demand more accurately and enhance the generalizability of these predictions across diverse regions.

Our work stands out by proposing a novel spatio-temporal prediction framework that synergistically integrates crowdsourced user-side KPIs with regulatory datasets. This innovative approach allows us to more accurately model and forecast spectrum demand by employing advanced temporal feature engineering, comprehensive correlation analyses, and state-of-the-art transfer learning techniques. With these innovations, we can effectively capture regional and temporal variations in spectrum utilization, moving beyond the static assumptions prevalent in traditional ITU models. Consequently, our framework offers policymakers and regulatory bodies a robust methodology for making informed decisions regarding spectrum management.
\section{Proposed Methodology}

Our proposed data-driven approach for estimating spectrum demand consists of five principal stages: (i) proxy definition and deployment, (ii) spatio-temporal data generation and preparation, (iii) feature (or KPI) engineering and correlation analysis, (iv) predictive model development and validation, and (v) transfer learning deployment. Fig.~\ref{fig1} provides a schematic overview of these components, clearly outlining the process flow and demonstrating how each element integrates into the overall methodology.
\subsection{Proxy Development}
To forecast spectrum demand accurately, we require a metric that reliably reflects actual network usage. We select aggregated deployed bandwidth—the total capacity allocated across all active sites in a region during a given time window—as this proxy. Because greater deployed bandwidth signals heavier user activity and higher network load, it offers a direct view of spectrum utilization. The metric is also practical: it can be retrieved from regulatory filings and operator reports, providing consistent, high-quality data. Therefore, adopting aggregated deployed bandwidth as our proxy gives the subsequent feature-engineering and modeling stages a robust, usage-based foundation.
\begin{figure}[!t]
    \centering
    \includegraphics[width=\columnwidth]{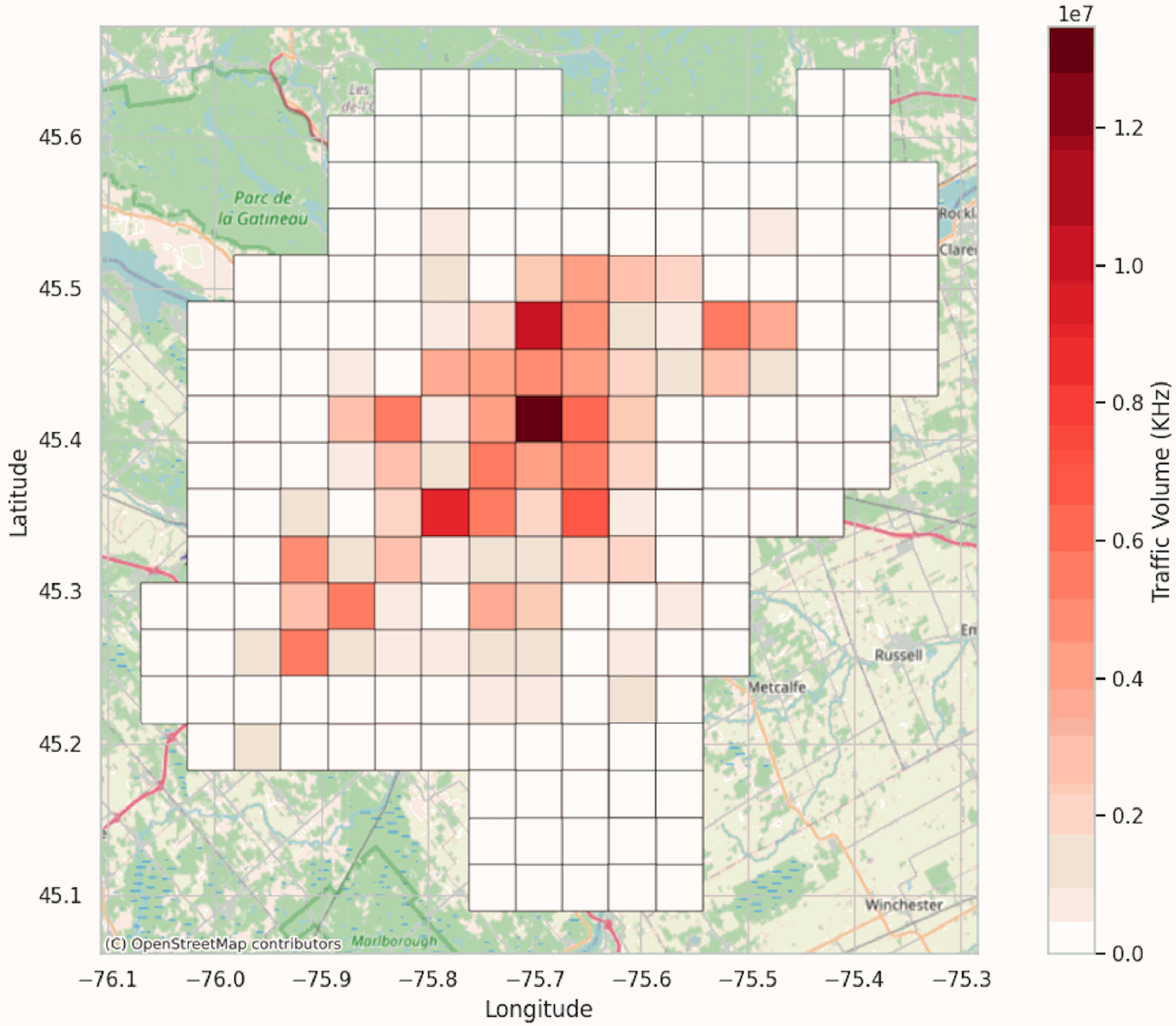}
    \caption{Spatio-temporal snapshot of a selected KPI in Ottawa, aggregated per geographic tile and rolling three-month window.}
    \label{fig2}
\end{figure}
\begin{figure}[!t]
    \centering
    \includegraphics[width=\columnwidth]{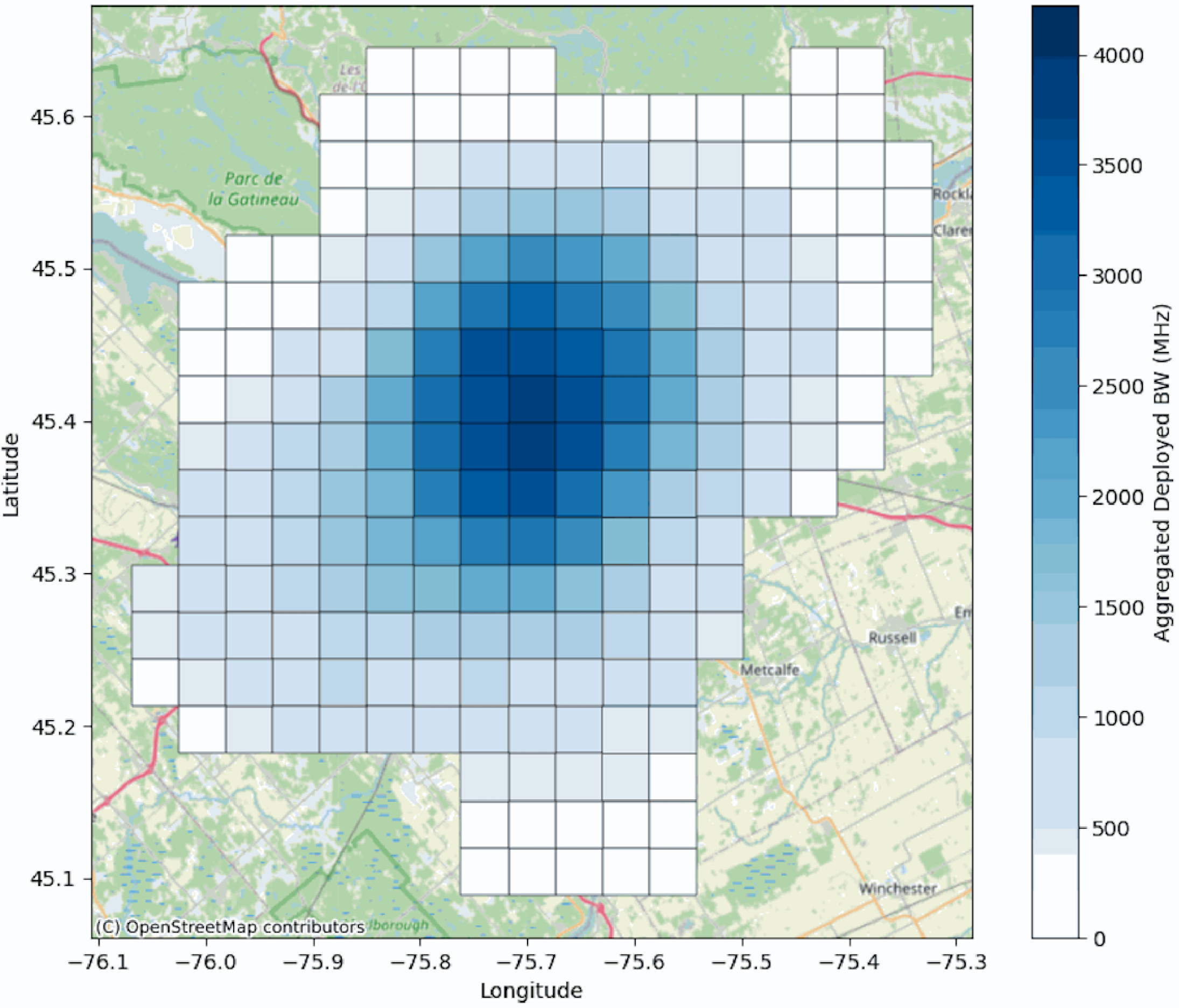}
    \caption{Spatio-temporal snapshot of the aggregated deployed-bandwidth proxy in Ottawa, aggregated per geographic tile and rolling three-month window.}
    \label{fig3}
\end{figure}
\subsection{Spatio-Temporal Data Generation}
Robust spectrum-demand forecasting begins with a data set that is both granular and consistent. We merge five years (2019–2023) of crowdsourced measurements, covering roughly 200 raw KPIs, with complementary regulatory records, and then reshape the combined data along two axes: space and time.

\begin{enumerate} \item \textbf{Spatial Aggregation:}
All observations are projected onto a common grid of geographical tiles sized to capture the contrast between dense urban cores, suburban belts, and sparsely populated rural areas. Within every tile–band pair, we average the raw samples, producing one record that summarises prevailing network conditions for that cell. This tiling yields a spatially coherent matrix in which each entry is directly comparable across regions, enabling fine-grained analyses of local spectrum-utilisation patterns.

\item \textbf{Temporal Segmentation:}
To maintain parity with the KPI calculations, the data are sliced into rolling three-month (quarterly) windows. This uniform cadence reveals seasonal cycles, growth trends, and behavioural shifts that affect demand. Aligning every KPI to the same quarterly timeline generates a set of consistent time series, simplifying lag creation, trend extraction, and transfer-learning across years and regions.

\item \textbf{Missing-Data and Outlier Treatment:}
A multi-step cleansing pipeline safeguards data integrity. Short gaps are closed with forward/backward interpolation, whereas longer gaps rely on moving-average smoothing or model-based imputation that borrows information from spatial neighbours and temporal lags. Outliers identified with inter-quartile-range and z-score tests—applied tile-by-tile and quarter-by-quarter—are winsorised or replaced using the same contextual models. These procedures ensure that the final KPI matrix is free of distortions and suitable for reliable prediction. \end{enumerate}

We create a high-fidelity spatio-temporal dataset that underpins all subsequent feature engineering and modelling stages by enforcing uniform spatial tiling, consistent quarterly segmentation, and rigorous quality control. Fig.~\ref{fig2} and~\ref{fig3} provide illustrative snapshots of the spatio-temporal maps for Ottawa. Fig.~\ref{fig2} visualizes the Traffic-Volume KPI, while Fig.~\ref{fig3} displays the aggregated deployed-bandwidth proxy. In both cases, values are averaged within each geographic tile and over a rolling three-month window.
\subsection{Feature Engineering and Correlation Analysis}

Effective spectrum-demand prediction depends on crafting a feature set that captures both the spatial–temporal dynamics of network activity and its causal link to our proxy, aggregated deployed bandwidth. The process unfolds in the following four tightly connected steps:
\begin{enumerate} \item \textbf{KPI Selection:}
A full audit of the crowdsourced traces yields a set of KPIs, each computed per tile, per band, and over rolling three-month windows from 2019 to 2022.
\emph{Traffic Volume} is the sum of average uplink and downlink throughputs;
\emph{Latency Ratio} divides the minimum by the mean one-way latency to capture delay stability;
\emph{Transmit-to-Receive Ratio} measures traffic asymmetry as bytes sent versus bytes received;
The \emph{Normalized Number of Connections} measures the connections per grid tile, normalized by the number of unique connected devices;
\emph{Signal Strength} reports radio-link quality;
\emph{Jitter Variability} is the difference between average and minimum jitter, reflecting packet-delay consistency; and
\emph{Sum Throughput} aggregates total bytes transmitted and received.
Together, these KPIs provide a balanced view of traffic load, latency behaviour, link symmetry, session activity, signal quality, delay variation, and aggregate volume.

\item \textbf{Lagged Features for Temporal Dependencies:}
Spectrum demand does not respond instantly to changes in network conditions; instead, it exhibits inertia that can span several months. To capture this behaviour, we augment every KPI with lagged counterparts offset by one and two quarters—the intervals that emerge as statistically significant in the proxy's autocorrelation and partial-autocorrelation analyses. Pairing each current KPI with its one and two quarter versions allows the model to separate short-term shocks (e.g., a sudden jump in Traffic Volume) from slower, structural trends (e.g., a steady rise in Sum Throughput). This ``Lagged-Regression'' architecture answers practical questions such as, ``How much extra spectrum will be required next quarter if uplink throughput spikes today, and will the effect persist six months down the line?'' Embedding this temporal memory substantially boosts forecast accuracy and improves generalisability across regions with different demand-evolution speeds.

\item \textbf{Correlation Analysis:}
With the expanded (current plus lagged) feature set in place, we conduct a two-stage correlation study. First, an inter-dataset check confirms alignment between the crowdsourced KPIs and the regulatory records of deployed bandwidth, ensuring both data sources narrate the same underlying demand story. Second, an intra-dataset analysis correlates each KPI, at lags zero, one, and two quarters, with the proxy. This step exposes leading indicators; for example, a strong correlation between the one-quarter-lagged Latency Ratio and future bandwidth deployments signals that rising latency today reliably foreshadows additional spectrum provisioning next quarter. Correlation magnitudes then guide feature ranking, weighting, and the removal of redundant predictors, yielding a concise yet temporally rich input set for the predictive models.

\item \textbf{Data Normalization:}
Finally, all features are standardized within each quarter to a zero mean and unit variance. This scaling removes biases caused by regional population density or network size differences, ensuring that the learning algorithms treat every feature and every region on an equal footing. \end{enumerate}

Collectively, these steps produce a coherent, well-scaled feature matrix that encapsulates the key spatial–temporal signals needed for high-fidelity spectrum-demand forecasting.
\subsection{Prediction Model and Validation}
The prediction framework leverages a hybrid modeling approach that balances interpretability with robust predictive performance, ensuring that both macro-level insights and micro-level predictive nuances are captured.

\textit{White-Box Models.}
Interpretability is a priority in our framework, and hence, we employ models such as Linear Regression and Lasso Regression. These models facilitate a clear understanding of the impact of individual KPIs on spectrum demand, enabling the identification of key drivers and providing transparency for policymakers and regulators. The explicit nature of white-box models allows for effective feature importance analysis and supports actionable recommendations based on quantifiable contributions.

\textit{Black-Box Models.}
To capture complex, non-linear relationships and interactions among various features, we utilize various non-linear models, including Random Forest and XGBoost, and LightGBM~\cite{ML-models}. These methods are particularly adept at accommodating heterogeneous network performance characteristics and fluctuating spectrum demand patterns that may vary spatially and temporally. The black-box models serve to uncover latent interactions that white-box models might overlook while complementing them through an ensemble strategy that enhances overall prediction accuracy.

\textit{Model Validation and Evaluation.}
The performance of both white-box and black-box models is rigorously evaluated using a suite of statistical metrics, including normalized root-mean-squared error (RMSE), actionable accuracy, and the coefficient of determination R-squared ($R^2$).
Furthermore, comparative analyses against traditional ITU benchmarks are conducted to validate the efficacy and practical relevance of our prediction framework. This comprehensive evaluation ensures that the model not only fits historical data accurately but also generalizes well to unseen data, thereby reaffirming its utility in diverse regional and temporal contexts.

This integrated use of both modeling approaches, supported by detailed spatio-temporal feature engineering and robust validation metrics, allows our framework to provide a nuanced and reliable estimation of spectrum demand that is both interpretable and highly predictive.
\begin{figure}[!t]
    \centering
    \hspace*{-3mm}
    \includegraphics[width=\columnwidth]{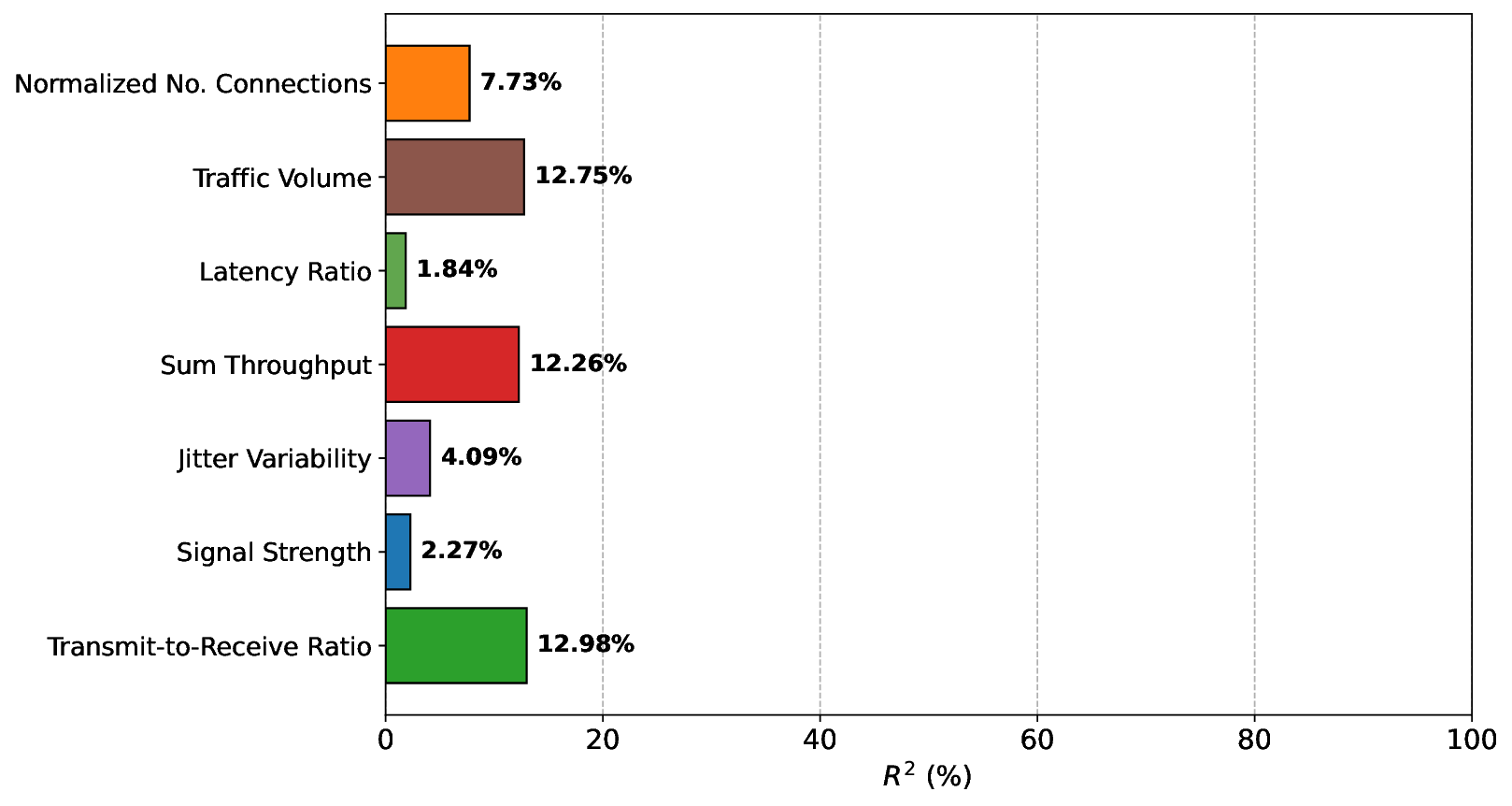}
    \caption{Correlation analysis between each KPI and proxy data, without considering the lagged KPI values, over 3‑month periods 2019‑2022, across all RF bands.}
    \label{fig:noLag}
\end{figure} 

\subsection{Transfer Learning Deployment}
Transfer learning is strategically implemented to enhance model generalizability and improve cross-regional prediction accuracy. In practice, a base predictive model is first trained on a source dataset, typically from regions with abundant historical data, which captures essential patterns in spectrum demand. Knowledge extracted from this source model is then transferred to target regions through a fine-tuning process, whereby model parameters are adjusted to account for localized variations in network characteristics and environmental factors. This methodology leverages domain adaptation techniques, ensuring that features learned in one geographical context are effectively re-calibrated for use in another, thereby reducing the computational overhead and data requirements typically associated with training from scratch.

Specifically, the transfer learning framework employs both layer freezing and gradual unfreezing strategies within deep learning architectures, such as stacked LSTM networks or convolutional neural networks (CNNs) for spatial data. By freezing the lower-level layers that capture universal patterns and fine-tuning only the upper layers, the approach maintains the integrity of previously learned generic features while adapting to region-specific peculiarities. This is particularly beneficial for regions with limited historical data, as it reduces the risk of overfitting and ensures that the model retains robustness across different network environments.

Furthermore, the integration of transfer learning enhances not only the predictive accuracy but also the model's ability to generalize to new, unseen data. It facilitates rapid deployment in diverse geographic areas by minimizing the need for extensive re-training, thereby accelerating the process of spectrum demand forecasting in response to evolving network conditions and policies.
\begin{figure}[!t]
    \centering
    \hspace*{-2mm}
    \includegraphics[width=\columnwidth]{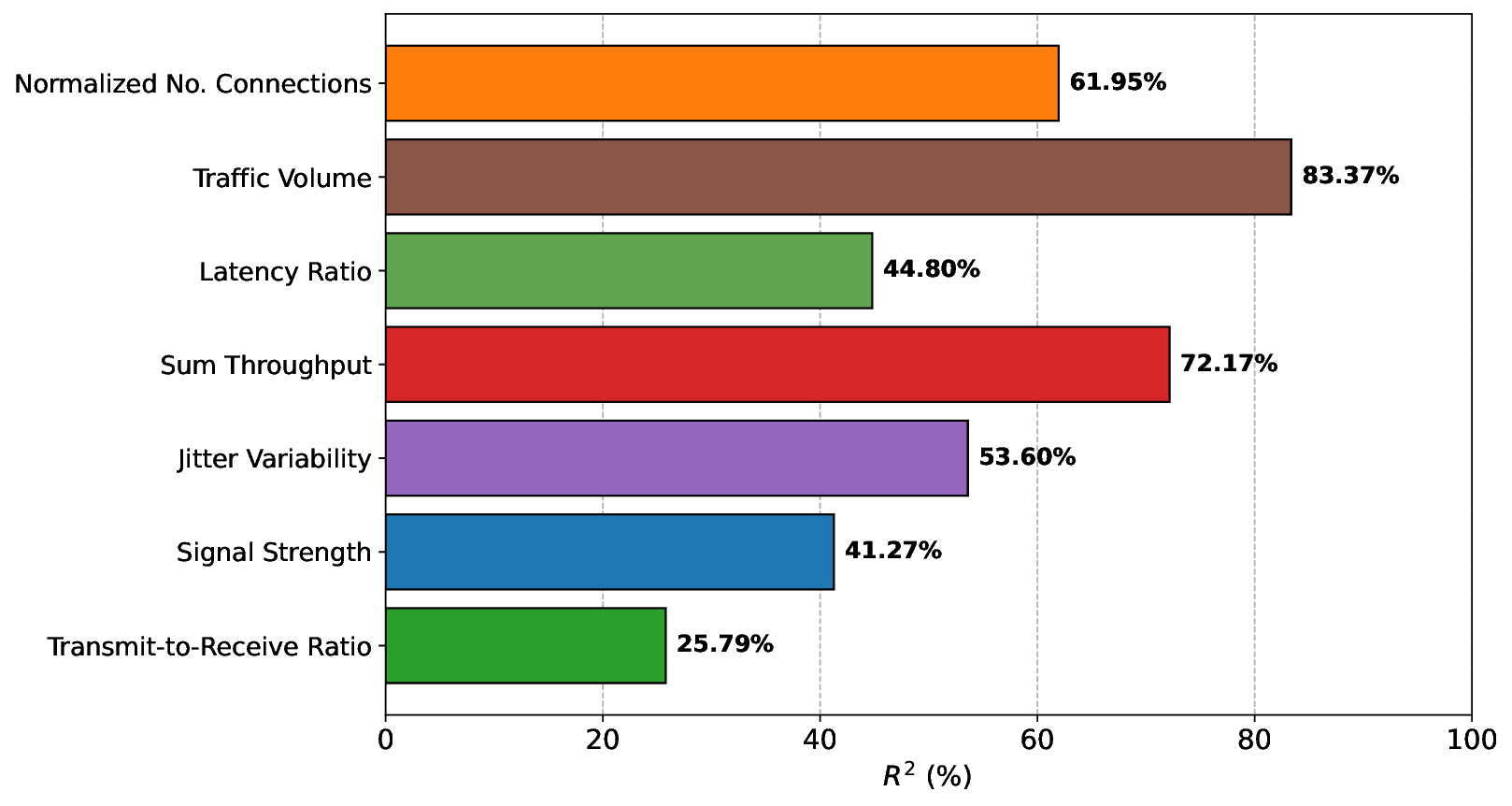}
    \caption{Correlation analysis considering both the current and lagged KPI values, over 3‑month periods 2019‑2022, across all RF bands.}
    \label{fig:withLag}
\end{figure} 
  \begin{figure}[!t]
     \centering
     \hspace*{-2mm}
     \includegraphics[width=\columnwidth]{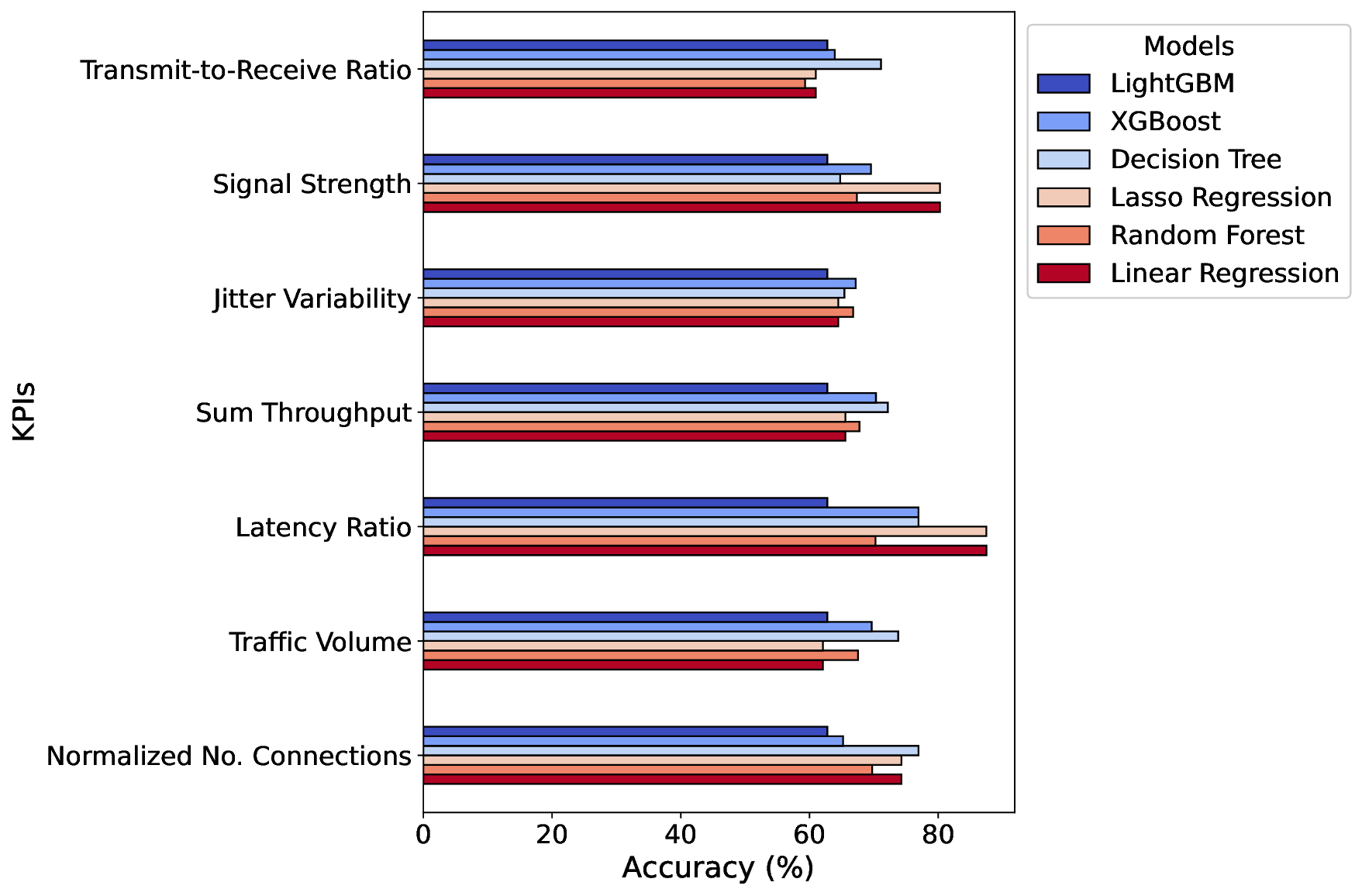}
     \caption{Prediction accuracy for selected KPIs across black-box and white-box models (2023-Ottawa).}
     \label{pred-models}
 \end{figure}
  \begin{figure}[!t]
     \centering
     \includegraphics[width=\columnwidth]{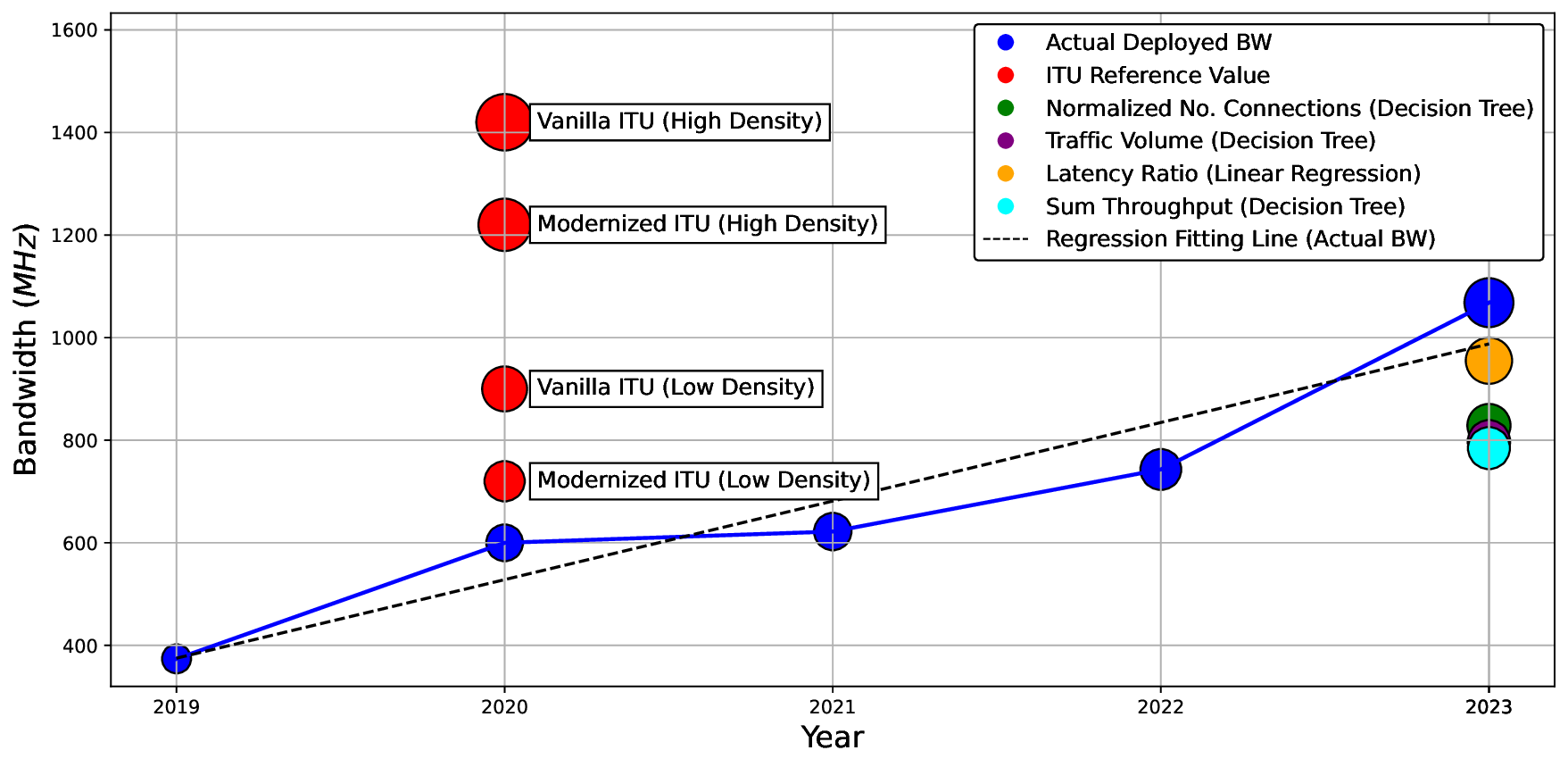}
     \caption{Comparing yearly actual deployed BW with KPI-based predictions (2023-Ottawa) and ITU benchmarks.}
     \label{fig:ITU_bench}
 \end{figure}

\section{Experimental Results}
This section evaluates the proposed methodology on a five-year (2019–2023) crowdsourced and regulatory dataset from Ottawa—a representative mid-tier Canadian market whose combination of a dense urban core and sparsely populated outskirts poses a stringent test of spatial-temporal generalization. Although the current analysis relies on 4G and 5G measurements, the framework is technology-agnostic and can be readily extended to forthcoming 6G networks.

Fig.~\ref{fig:noLag} shows the correlations between each KPI and the deployed-bandwidth proxy when only current‐quarter values are used. Fig.~\ref{fig:withLag} repeats the analysis after adding +1 quarter lags. Incorporating lagged terms raises correlation magnitudes for every metric, most notably Traffic Volume from $12.75\%$ to $83.37\%$ at +1 quarter. The result confirms that spectrum provisioning reacts with a delay to changes in user-side performance, validating the Lagged-Regression feature set adopted in the predictive models. 

Fig.~\ref{pred-models} benchmarks several white- and black-box models, including Linear Regression, Lasso, Decision Tree, Random Forest, LightGBM, and XGBoost, using the complete KPI plus lag feature set. Linear and Lasso regressions yield the highest performance, achieving a normalized RMSE that translates to an accuracy of roughly $86\%$. XGBoost follows, lagging by about ten percentage points, while the remaining tree-based models trail further. Because the two regression models are also fully interpretable, they provide clear policy guidance, whereas ensemble methods offer additional capacity to model nonlinear effects. These results support a hybrid strategy: use transparent linear models for regulatory insight and deploy tree-based learners when maximum predictive accuracy is required.

Fig.~\ref{fig:ITU_bench} compares our annual spectrum‑demand forecasts (2019–2023) with the four ITU reference points issued for 2020, including Vanilla ITU (high/low density) and Modernized ITU (high/low density). We developed the Modernized ITU model by making the assumption that all traffic on the network was packet-switched, rather than a mix of circuit-switched and packet-switched as specified in the original model. Across all study areas, the ITU benchmarks consistently over‑predict demand. In secondary markets, realised bandwidth remains $20$-$40\%$ below even Modernized ITU (low‑density); in the Greater Toronto Area, it surpasses that modernized value yet still trails Vanilla ITU (low‑density) by roughly $15\%$. These deviations highlight the inability of a single‑year, parameter‑fixed ITU model to accommodate regional traffic heterogeneity and deployment strategies. Our spatio‑temporal framework, driven by seven engineered KPIs and their $+1$ quarter lags, achieves markedly tighter alignment with observed deployments. Ordinary least‑squares and Lasso regressions yield accuracy (normalized RMSE) up to $0.85$ across regions, demonstrating their capacity to model both immediate and lag‑induced effects. The empirical evidence thus confirms that granular, KPI‑based models provide a more accurate and operationally actionable representation of spectrum demand than static ITU projections.

\begin{figure*}[!t]
    \centering
    \hspace*{-3mm}
    \includegraphics[width=1\linewidth]{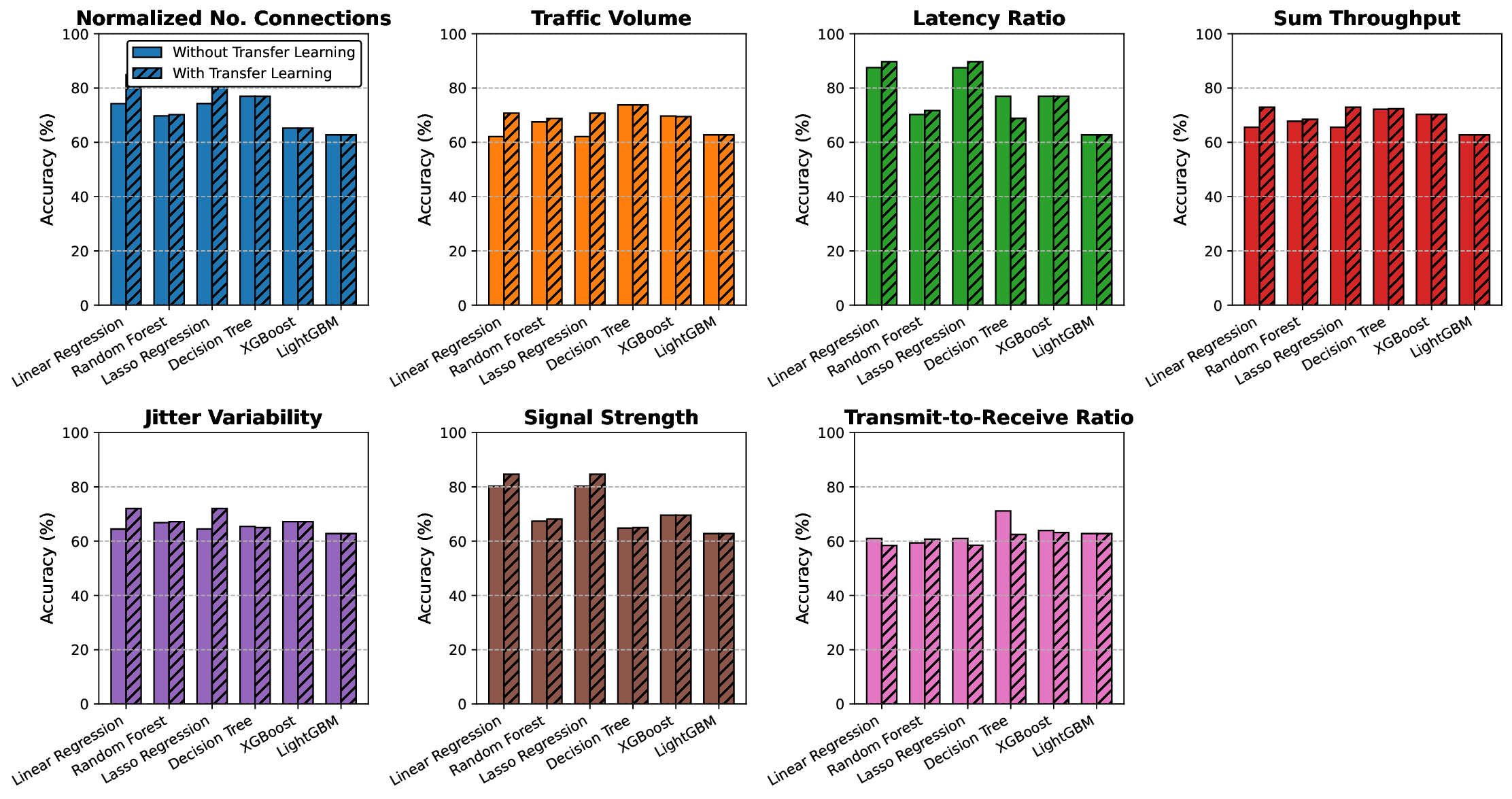}
    \caption{Accuracy performance of KPI-based prediction models using transfer learning and non-transfer learning approaches.}
    \label{transfer}
\end{figure*}

Fig.~\ref{transfer} compares the prediction accuracy with and without transfer learning. We first train the regression model on Toronto's data-rich KPI matrix and proxy values, then fine-tune it for Ottawa using its available dataset while passing Toronto-derived proxy estimates as an additional feature. Compared with a model trained from scratch on Ottawa data, this transfer-learning configuration reduces the normalized RMSE by up to $15\%$, yielding a commensurate gain in accuracy. The improvement confirms that KPI–demand relationships inferred in a dense urban environment can be ported, with minimal adaptation, to a medium-density market, offering a practical pathway for rapid, reliable forecasting in regions that lack extensive historical records.

\section{Conclusion and Future Work}

Leveraging crowdsourced, user-side KPIs, this work delivers a spatio-temporal spectrum-demand model that is both finer-grained and more adaptive than static ITU benchmarks. This framework is technology-agnostic and can be readily
extended to forthcoming 6G networks. By pairing each KPI with quarterly lags and applying transfer learning, the framework cuts normalized RMSE by up to $15\%$ and maintains accuracy across dense and sparse markets—performance unattainable with conventional, parameter-fixed estimates. Remaining challenges include data sparsity in very low-population tiles, potential sampling bias in crowdsourced inputs, and the need for periodic retraining as deployment patterns evolve. Addressing these limitations will further tighten forecast fidelity. Future work will (i) incorporate richer KPIs (such as application-level traffic), (ii) extend transfer-learning pipelines to multi-country datasets, and (iii) investigate attention-based transformers and meta-learning for continual, self-adapting forecasts. 


\bibliographystyle{IEEEtran}

\end{document}